\documentclass[conference]{IEEEtran}
\IEEEoverridecommandlockouts
\usepackage{cite}
\usepackage{amsmath,amssymb,amsfonts}
\usepackage{algorithmic}
\usepackage{xspace}

\usepackage{textcomp}
\usepackage{xcolor}
\usepackage{enumitem}
\usepackage{booktabs} 
\usepackage{multirow}
\usepackage{caption}
\usepackage{colortbl} 

\definecolor{lavender}{HTML}{D9D9FF}    
\definecolor{lightgrayrow}{HTML}{ECECEC} 

\usepackage{graphicx}

\newcolumntype{C}[1]{>{\centering\arraybackslash}m{#1}}

\usepackage{booktabs}      
\usepackage{multirow}

\usepackage{arydshln}      
\usepackage[table]{xcolor} 

\definecolor{firstcolor}{RGB}{255,220,220}   
\definecolor{secondcolor}{RGB}{220,235,255}  
\usepackage{booktabs}   
\usepackage{multirow}
\usepackage{array}

\usepackage{colortbl}
\usepackage{arydshln}   

\newcolumntype{M}{>{\raggedright\arraybackslash}m{3.2cm}}

\newcommand{\GrayThickDash}{%
  \noalign{\global\setlength{\arrayrulewidth}{0.8pt}}%
  \arrayrulecolor{black!35}\hdashline[8pt/3pt]\arrayrulecolor{black}%
  \noalign{\global\setlength{\arrayrulewidth}{0.4pt}}%
}

\def\BibTeX{{\rm B\kern-.05em{\sc i\kern-.025em b}\kern-.08em
T\kern-.1667em\lower.7ex\hbox{E}\kern-.125emX}}

\makeatletter

\DeclareRobustCommand\onedot{\futurelet\@let@token\@onedot}
\def\@onedot{\ifx\@let@token.\else.\null\fi\xspace}

\def\etal{\emph{et al}\onedot}
\makeatother
\begin{document}

\title{Evidence Packing for Cross-Domain Image Deepfake Detection with LVLMs}

\author{Yuxin Liu$^{1,3}$, Fei Wang$^{2,3,*}$, Kun Li$^{4}$, Yiqi Nie$^{1,3}$, Junjie Chen$^{2,3}$,  Zhangling Duan$^{3,*}$, Zhaohong Jia$^1$\thanks{* Corresponding Author: Fei Wang, Zhangling Duan}\\
\small{$^1$ Anhui University}~~
\small{$^2$ Hefei University of Technology}\\
\small{$^3$ IAI, Hefei Comprehensive National Science Center}~~
\small{$^4$ United Arab Emirates University}}
\maketitle

\begin{abstract}
Image Deepfake Detection (IDD) separates manipulated images from authentic ones by spotting artifacts of synthesis or tampering. Although large vision-language models (LVLMs) offer strong image understanding, adapting them to IDD often demands costly fine-tuning and generalizes poorly to diverse, evolving manipulations. We propose the Semantic Consistent Evidence Pack (SCEP), a training-free LVLM framework that replaces whole-image inference with evidence-driven reasoning. SCEP mines a compact set of suspicious patch tokens that best reveal manipulation cues.
It uses the vision encoder’s CLS token as a global reference, clusters patch features into coherent groups, and scores patches with a fused metric combining CLS-guided semantic mismatch with frequency- and noise-based anomalies.
To cover dispersed traces and avoid redundancy, SCEP samples a few high-confidence patches per cluster and applies grid-based NMS, producing an evidence pack that conditions a frozen LVLM for prediction.
Experiments on diverse benchmarks show SCEP outperforms strong baselines without LVLM fine-tuning.
\end{abstract}


\section{Introduction}
\label{sec:intro}
Image Deepfake Detection (IDD) aims to identify manipulated images by analyzing visual cues that reveal synthetic generation or deliberate tampering, playing a critical role in improving the reliability and trustworthiness of modern information systems~\cite{pei2403deepfake}.
The central challenge in IDD is to capture subtle, localized manipulation traces, such as blending artifacts at boundaries or biases and inconsistencies in low-level image statistics~\cite{yermakov2025deepfake}.
Recent studies~\cite{zhang2024common}~\cite{nguyen2025prpo} have increasingly leveraged large vision-language models (LVLMs), reformulating IDD as a Visual Question Answering (VQA)-style recognition problem and fine-tuning pretrained backbones to enhance robustness under real-world conditions.
Zhang~\etal~\cite{zhang2024common} formulate IDD as a VQA task, modeling human intuition by generating textual rationales that articulate common-sense cues for classifying an image as real or fake.
Nguyen~\etal~\cite{nguyen2025prpo} modify LLaVA with a texture-biased CLIP visual encoder and further encourage the model’s test-time reasoning to remain grounded in visual evidence, improving the reliability of IDD.
However, such methods typically rely on high-quality supervision, including carefully crafted question templates and human-authored textual rationales.
Although LVLMs demonstrate strong image understanding capabilities, fine-tuning them for deepfake image detection often incurs substantial computational overhead and still exhibits limited generalization across diverse manipulation types~\cite{Funtuned_qppoch}.

Image key token selection aims to identify a compact subset of informative visual tokens that preserves salient image content while reducing redundancy and computational cost for downstream  reasoning~\cite{koner2024lookupvit}.
To this end, our research focuses on the following challenges:
(i) Training LVLMs typically requires substantial computational resources, and applying IDD to training-free LVLMs faces the difficulty of ensuring robust generalization across diverse forgery types and scenarios.
(ii) Generic token selection strategies, largely optimized for semantics or efficiency, often remove fine-grained manipulation cues in training-free LVLMs for IDD, diluting evidence and degrading reliability.
(iii) Token selection is sensitive to an image’s low-level statistics, which can distract it from forgery-relevant cues.
Integrating forgery-aware cues and low-level statistical priors into token selection can better surface manipulation-relevant tokens, improving the robustness of training-free LVLMs for IDD.
\begin{figure}[t!]
\begin{center}
\includegraphics[width=1\linewidth]{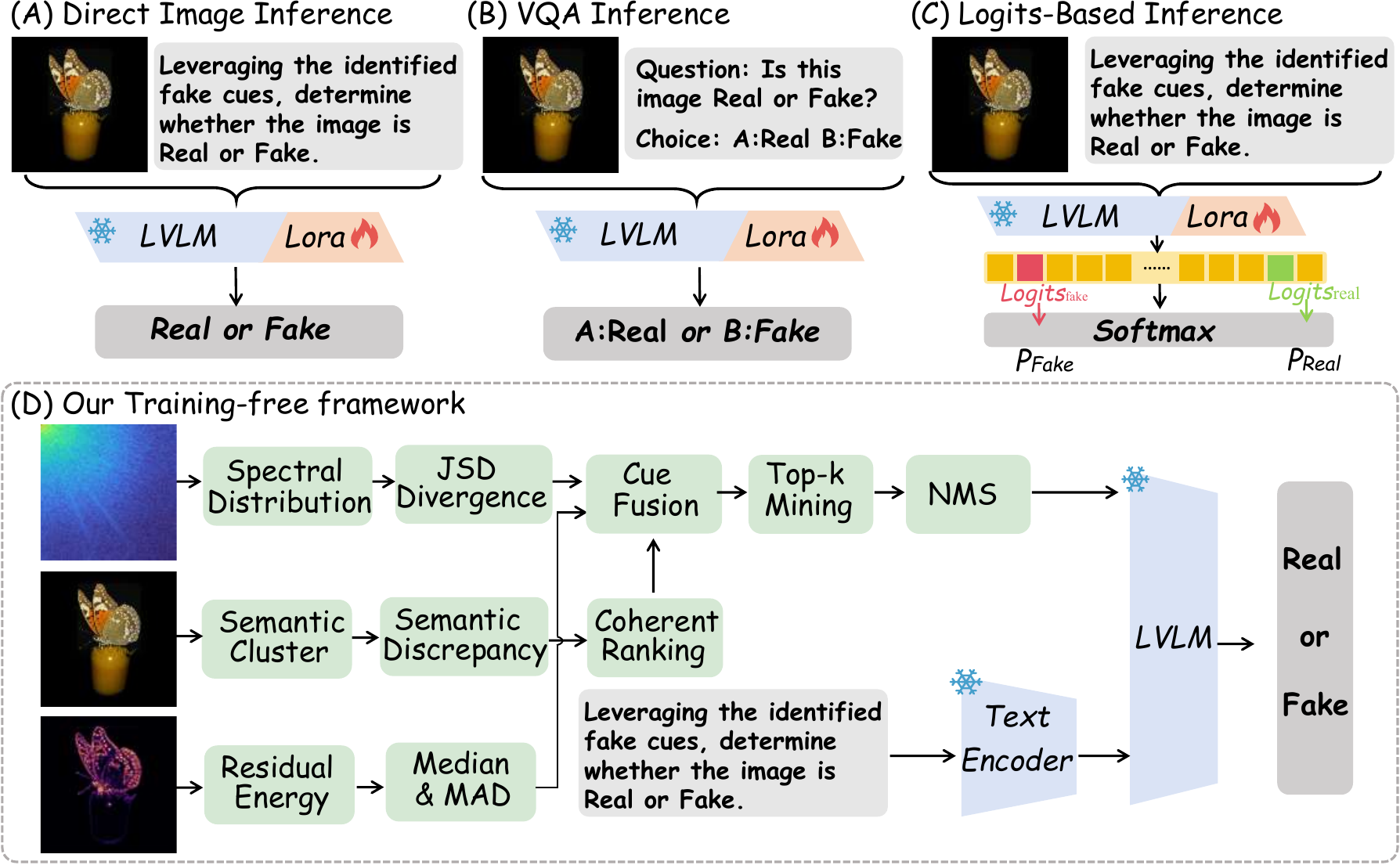}
\vspace{-1.0em}
\caption{
\textbf{
Comparison of LVLM inference paradigms for IDD.}
(A) Direct inference uses the full image.
(B) VQA-style inference formulates IDD.
(C) Logits-based inference uses token probabilities.
(D) A training-free, evidence-driven framework that selects compact evidence tokens with frequency/noise cues and grid-based NMS for LVLM prediction.
}
\label{fig:overall}
\end{center}
\vspace{-0.8em}
\end{figure}
Motivated by these challenges, we propose \textbf{Semantic Consistent
Evidence Pack (SCEP)}, a training-free deepfake cue mining framework that selectively isolates forgery-relevant patch tokens and suppresses irrelevant semantic content.
(1) We feed the input image into a frozen vision encoder to obtain the CLS token and patch embeddings. 
To enforce semantic consistency, we cluster patch features into coherent groups and perform evidence mining and suspect-control contrast only within each group, encouraging the LVLM to focus on manipulation cues over semantic content.
(2) Within each semantic cluster, SCEP assigns every valid patch token a suspiciousness score by combining a CLS-guided semantic discrepancy with patch-level frequency and noise anomalies.
It retains a small set of top-ranked tokens per cluster and applies grid-based NMS to suppress spatially redundant candidates.
The final verdict is obtained by aggregating evidence across clusters and prompting a frozen LVLM with the assembled evidence pack.
As shown in Fig.~\ref{fig:overall}, existing IDD methods span supervised LVLM fine-tuning, training-free whole-image inference with frozen LVLMs, and VQA- or logits-based formulations, often relying on carefully designed question templates and human-authored rationales.
In contrast, our framework shifts the inference paradigm from reasoning over the entire image to evidence-driven reasoning.
The contributions of this work are summarized as follows:
\begin{itemize}[noitemsep, topsep=0pt, leftmargin=10 pt]

\item We propose SCEP, a training-free LVLM framework that replaces whole-image inference with weighted evidence reasoning, reducing redundant tokens and improving sensitivity to subtle forgery cues.

\item SCEP anchors semantic clustering with the frozen CLS token and fuses CLS-supervised semantic discrepancy with low-level frequency and noise anomalies into a unified cluster-wise suspiciousness score.

\item Top-ranked tokens are selected within each cluster and refined with grid-based NMS to reduce semantic and spatial redundancy, yielding a compact, spatially diverse evidence set with cross-cluster coverage.

\item Extensive experiments demonstrate that SCEP substantially improves IDD by supplying more informative and less redundant evidence for LVLM inference.
\end{itemize}
\begin{figure*}[t!]
\begin{center}
\includegraphics[width=1.0\linewidth]{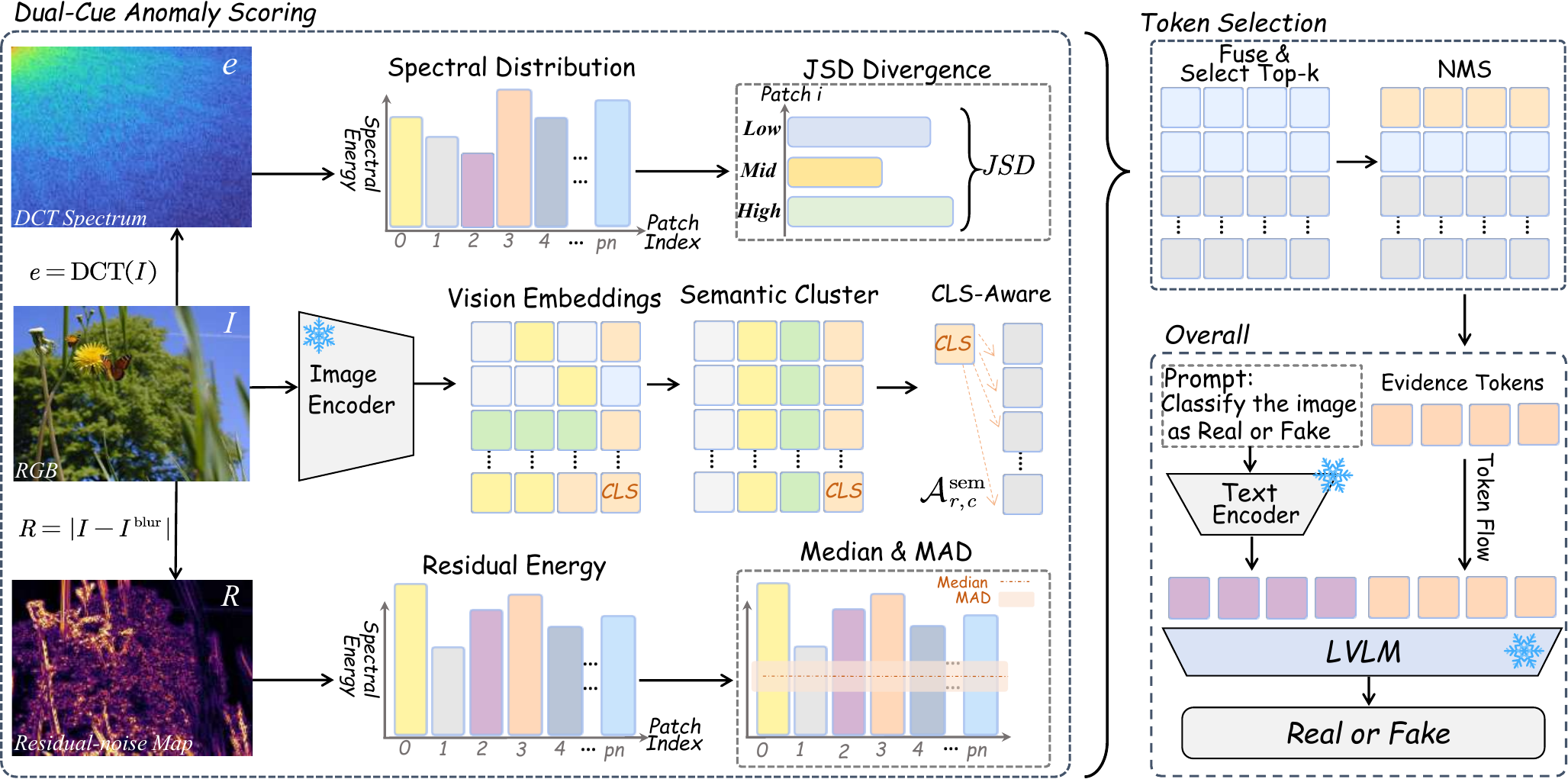}
\vspace{-1.5em}
\caption{
\textbf{Overview of SCEP for training-free IDD.}
(1) Dual-cue patch anomalies are scored in semantic space: frequency is measured by JSD between each patch’s DCT spectrum, while noise is quantified from residuals via the median and MAD.
(2) Patch embeddings are clustered around the CLS anchor to select top-k evidence tokens, and grid-based NMS removes redundant ones.
The resulting evidence pack conditions a frozen LVLM to predict Real or Fake.}
\label{fig:method}
\end{center}
\vspace{-1.5em}
\end{figure*}

\section{Methodology}
\subsection{Overview}
This work focuses on IDD by leveraging LVLMs without task-specific fine-tuning.
We cast IDD as a binary decision problem on multimodal inputs.
Each instance contains a visual input $I \in \mathbb{R}^{H \times W \times 3}$, and the model outputs a label indicating whether the content is manipulated or authentic.
In the standard inference setting, we consider a frozen LVLM $\mathcal{M}(\cdot)$, which takes an input image $I$ together with a prompt  $\mathcal{P}\in \mathbb{R}^{L}$ and outputs a decision $\mathcal{F}$ as
\begin{equation}
\mathcal{F} = \mathcal{M}(I, \mathcal{P}).
\end{equation}

However, directly performing inference with a frozen LVLM often fails to precisely capture subtle forgery cues in $I$, while producing a large number of redundant visual tokens during decoding.
To address this limitation, we propose the \textbf{Semantic Consistent Evidence Pack (SCEP)}, a training-free framework that extracts the vision encoder class token CLS as a global semantic anchor, clusters patch embeddings to enforce semantic consistency, and ranks suspicious tokens by fusing CLS-supervised semantic discrepancy with frequency and noise anomaly cues, thereby yielding a weighted evidence pack for LVLM inference, as illustrated in Fig.~\ref{fig:method}.
(1) We feed the input image $I$ into a frozen vision encoder to extract the class token $t_{[CLS]}$ and patch embeddings $\{\mathbf{t}_{r,c}\}_{r=1}^{g_h}{}_{c=1}^{g_w}$, where $g_w$ and $g_h$ denote the number of patch tokens along the vertical and horizontal axes.
Next, clustering is performed in the patch embedding space to obtain $K$ semantic groups $\{\mathrm{C}_k\}_{k=1}^{K}$, such that evidence mining and suspect control contrast are confined to semantically consistent regions, allowing the LVLM to focus on manipulation cues rather than semantic content.
(2) Within each semantic cluster ${\mathrm{{C}_k}}$, SCEP scores each valid token $\mathbf{t}_{r,c}\in{\mathrm{C_k}}$ using CLS-supervised semantic discrepancy, which provides a global semantic prior for identifying semantically inconsistent patches.
In parallel, SCEP computes low-level anomaly cues at the patch scale, including a frequency score $\mathcal{A}^{\mathrm{freq}}_{r,c}$ that captures deviations in the patch DCT band energy distribution and a noise score and a noisy score $\mathcal{A}^{\mathrm{noise}}_{r,c}$ derived from inconsistencies in high-frequency residual statistics.
Tokens are then ranked within ${\mathrm{{C}_k}}$, and a preliminary candidate set $\mathrm{\hat{S}_{k}}\subseteq {\mathrm{{C}_k}}$ is obtained by selecting the top-$k_1$ tokens.
To reduce spatial redundancy, we further apply a grid-based non-maximum suppression operator $\mathrm{NMS(\cdot)}$, which discards candidates within an $\ell_2$ distance of any higher scored token, yielding the final evidence set $\mathrm{S_k}\subseteq\mathrm{\hat{S}_{k}}$.
The final prediction $\hat{\mathcal{F}}$ is obtained by aggregating all evidence pairs across clusters as $\mathrm{S_k}$ and encoding the prompt into a token sequence $\hat{\mathrm{P}}\in\mathbb{R}^{L}$.
The LVLM $\mathcal{M}(\cdot)$ performs inference with $\mathrm{S_k}$ and $\hat{\mathrm{P}}$:
\begin{equation}
\hat{\mathcal{F}} = \mathcal{M}(\mathrm{S_k}, \hat{\mathrm{P}}).
\end{equation}

\subsection{Semantic Clustering and CLS-Supervised Scoring}
The primary objective of semantic space processing is to isolate and highlight manipulation cues that deviate from the global image context, enabling more focused and accurate deepfake detection~\cite{guan2022delving}.
To this end, we adopt a two-step pipeline that performs semantic clustering over patch embeddings and then scores semantic discrepancies under CLS supervision. 
Concretely, given an input image $I$, we feed it into a frozen vision encoder to obtain the class token $\mathbf{t}_{CLS}$ and the patch embeddings $\{\mathbf{t}_{r,c}\}_{r=1}^{g_h}{}_{c=1}^{g_w}$.
Inspired by prior work on spherical clustering~\cite{hornik2012spherical}, spherical $k$-means is applied to cluster patch embeddings by angular similarity, yielding semantically coherent regions that group visually similar patches.

For spherical $k$-means, we assign each normalized embedding to the centroid with the highest cosine similarity:
\begin{equation}
z_{r,c}=\arg\max_{j\in\{1,\dots,K\}} \tilde{\mathbf{t}}_{r,c}^{\top}\boldsymbol{\mu}_j,
\end{equation}
\begin{equation}
\mathcal{C}_k=\{\mathbf{t}_{r,c}\mid z_{r,c}=k\},
\end{equation}
where $\tilde{\mathbf{t}}_{r,c}=\mathbf{t}_{r,c}/\|\mathbf{t}_{r,c}\|_2$ is the normalized embedding and $\boldsymbol{\mu}_j$ is the centroid of cluster $j$.

In IDD, manipulation cues are subtle and localized, and naively ranking all patch tokens is often dominated by semantically irrelevant regions.
To prioritize regions whose local semantics diverge from the overall image content, we measure token-level inconsistency with respect to the global context encoded by  $\mathbf{t}_{[CLS]}$ and rank tokens within each cluster $\mathcal{C}_k$.
Concretely, for each toke $\mathbf{t}_{r,c}\in\mathcal{C}_k$, we compute a CLS-supervised semantic discrepancy $\mathcal{A}^{\mathrm{sem}}_{r,c}$ by converting cosine similarity into a distance:
\begin{equation}
\mathcal{A}^{\mathrm{sem}}_{r,c}
=
1-\frac{\mathbf{t}_{r,c}^{\top}\mathbf{t}_{[CLS]}}{\|\mathbf{t}_{r,c}\|_2\,\|\mathbf{t}_{[CLS]}\|_2}.
\end{equation}

\subsection{Frequency Anomaly Cue via DCT Band Discrepancy}
To better capture subtle manipulation cues that are often missed by conventional analyses, two complementary anomaly cues are introduced, namely frequency anomalies and noise anomalies~\cite{kashiani2025freqdebias}~\cite{wang2024frequency}~\cite{li2025mmad}.
They probe irregularities in the frequency domain and residual noise statistics, respectively, thereby exposing manipulation artifacts that are difficult to detect from pixel-space evidence alone.

For each patch $p_{r,c}$, we apply a 2D DCT to obtain its coefficient map $\mathbf{F}_{r,c}$ from which we summarize the patch spectrum by aggregating band-wise DCT energy over $K$ pre-defined frequency bands \(\{\mathcal{B}_k\}_{k=1}^{K}\).
Band-wise normalization then yields the patch-level frequency distribution vector $\mathbf{q}_{r,c}\in\mathbb{R}^{K}$, where each component is computed as
\begin{equation}
q_{r,c}^{(k)}=\frac{e_{r,c}^{(k)}}{\sum_{j=1}^{K}e_{r,c}^{(j)}} .
\end{equation}

To compare local spectra against an image-level baseline in IDD, the reference frequency profile is computed by averaging the patch-wise distributions $\bar{{\mathbf{q}}}$:
\begin{equation}
\bar{\mathbf{q}}=\frac{1}{g_h g_w}\sum_{r=1}^{g_h}\sum_{c=1}^{g_w}\mathbf{q}_{r,c},
\end{equation}
where $g_w$ and $g_h$ denote the number of patch tokens along the vertical and horizontal axes.

Following recent IDD studies that measure distributional discrepancies via Jensen-Shannon divergence $\mathrm{JSD}(\cdot)$~\cite{liu2024bottom}, each patch is assigned a frequency anomaly score $\mathcal{A}^{\mathrm{freq}}_{r,c}$ by comparing its local spectrum distribution to the image-level profile
\begin{equation}
\mathcal{A}^{\mathrm{freq}}_{r,c}
=\mathrm{JSD}\!\left(\mathbf{q}_{r,c}\,\Vert\,\bar{\mathbf{q}}\right).
\end{equation}

\begin{table*}[t!]

\centering
\setlength{\tabcolsep}{3pt}
\renewcommand{\arraystretch}{1.08} 

\caption{\textbf{Performance on DFBench across AI-generated subsets.} The table compares multiple frozen LVLM backbones and token-pruning baselines, A-ViT and DynamicViT. Rows shaded in \colorbox{lavender}{\strut\textcolor{black}{lavender}} indicate token-pruning baselines, while the row shaded in \colorbox{lightgrayrow}{\strut\textcolor{black}{light gray}} highlights SCEP.}

\label{tab:merged}
\vspace{-0.6em}

\resizebox{\textwidth}{!}{
\begin{tabular}{@{} C{2.2cm} *{14}{c} @{}}
\toprule

\multicolumn{1}{c}{\textbf{Datasets}}
& \multicolumn{2}{c}{\textbf{Playground}~\cite{Playgroun}}
& \multicolumn{2}{c}{\textbf{SD3.5 Large}~\cite{SD5LARGE}}
& \multicolumn{2}{c}{\textbf{PixArt-Sigma}~\cite{PixArt-sigma}}
& \multicolumn{2}{c}{\textbf{Infinity}~\cite{Infinity}}
& \multicolumn{2}{c}{\textbf{Kandinsky}-3~\cite{Kandinsky}}
& \multicolumn{2}{c}{\textbf{Flux Schnell}~\cite{Flickr8k}}
& \multicolumn{2}{c}{\textbf{Kolors}~\cite{kolors}} \\
\cmidrule(lr){2-3}\cmidrule(lr){4-5}\cmidrule(lr){6-7}\cmidrule(lr){8-9}%
\cmidrule(lr){10-11}\cmidrule(lr){12-13}\cmidrule(lr){14-15}
\multicolumn{1}{c}{\textbf{Methods / Metrics}}
& \multicolumn{1}{c}{\textbf{Acc\%}} & \multicolumn{1}{c}{\textbf{F1}}
& \multicolumn{1}{c}{\textbf{Acc\%}} & \multicolumn{1}{c}{\textbf{F1}}
& \multicolumn{1}{c}{\textbf{Acc\%}} & \multicolumn{1}{c}{\textbf{F1}}
& \multicolumn{1}{c}{\textbf{Acc\%}} & \multicolumn{1}{c}{\textbf{F1}}
& \multicolumn{1}{c}{\textbf{Acc\%}} & \multicolumn{1}{c}{\textbf{F1}}
& \multicolumn{1}{c}{\textbf{Acc\%}} & \multicolumn{1}{c}{\textbf{F1}}
& \multicolumn{1}{c}{\textbf{Acc\%}} & \multicolumn{1}{c}{\textbf{F1}} \\
\midrule

Kimi-VL  &10.53&0.36&9.72&0.24&15.22&0.25&5.65&0.14&15.98&0.33&3.32&0.08&6.54&0.33 \\
MiniCPM-V
&26.43&0.39&3.56&0.07&32.33&0.42&8.67&0.12&21.22&0.23&2.32&0.29&13.76&0.27 \\
Janus-Pro
&10.33&0.07&3.48&0.08&2.27&0.31&2.23&0.03&0.34&0.17&0.31&0.27&0.36&0.12 \\
Owl2.1
&12.43&0.18&2.03&0.04&18.93&0.29&6.43&0.05&17.83&0.21&21.86&0.16&15.73&0.18\\
Gemma-3
&41.21&0.54&32.33&0.26&48.66&0.64&22.63&0.29&30.24&0.43&23.69&0.31&21.78&0.37 \\
\rowcolor{blue!15}+A-ViT ~\cite{avit}&42.21&0.56&33.43&0.27&48.73&0.67&26.23&0.33&32.24&0.46&24.12&0.31&23.54&0.39\\
\rowcolor{blue!15}+DynamiViT ~\cite{dynamicvit}&42.76&0.57&35.76&0.32&48.86&0.69&27.12&0.35&37.24&0.46&26.87&0.31&23.65&0.43\\
\rowcolor{gray!15} \textbf{+SCEP} (Ours)
&\textbf{43.33}&\textbf{0.61}&\textbf{36.41}&\textbf{0.34}&\textbf{49.12}&\textbf{0.73}&\textbf{28.67}&\textbf{0.41}&\textbf{39.87}&\textbf{0.50}&\textbf{31.66}&\textbf{0.34}&\textbf{30.21}&\textbf{0.45}\\

\GrayThickDash

\multicolumn{1}{c}{\textbf{Datasets}}
& \multicolumn{2}{c}{\textbf{SD3 Medium}~\cite{SD5LARGE}}
& \multicolumn{2}{c}{\textbf{Flux Dev}~\cite{Flickr8k}}
& \multicolumn{2}{c}{\textbf{NOVA}~\cite{nova}}
& \multicolumn{2}{c}{\textbf{LaVi-Bridge}~\cite{llavabridge}}
& \multicolumn{2}{c}{\textbf{Janus}~\cite{Janus}}
& \multicolumn{2}{c}{\textbf{Real Source}}
& \multicolumn{2}{c}{\textbf{Overall}} \\
\cmidrule(lr){2-3}\cmidrule(lr){4-5}\cmidrule(lr){6-7}\cmidrule(lr){8-9}%
\cmidrule(lr){10-11}\cmidrule(lr){12-13}\cmidrule(lr){14-15}
\multicolumn{1}{c}{\textbf{Methods / Metrics}}
& \multicolumn{1}{c}{\textbf{Acc\%}} & \multicolumn{1}{c}{\textbf{F1}}
& \multicolumn{1}{c}{\textbf{Acc\%}} & \multicolumn{1}{c}{\textbf{F1}}
& \multicolumn{1}{c}{\textbf{Acc\%}} & \multicolumn{1}{c}{\textbf{F1}}
& \multicolumn{1}{c}{\textbf{Acc\%}} & \multicolumn{1}{c}{\textbf{F1}}
& \multicolumn{1}{c}{\textbf{Acc\%}} & \multicolumn{1}{c}{\textbf{F1}}
& \multicolumn{1}{c}{\textbf{Acc\%}} & \multicolumn{1}{c}{\textbf{F1}}
& \multicolumn{1}{c}{\textbf{Acc\%}} & \multicolumn{1}{c}{\textbf{F1}} \\
\midrule

Kimi-VL&9.32&0.14&18.00&0.33&16.31&0.33&22.03&0.28&17.01&0.33&96.02&0.68&18.89&0.29\\
MiniCPM-V&30.12&0.28&5.55&0.18&30.98&0.37&10.00&0.22&18.99&0.18&95.56&0.51&23.03&0.27\\
Janus-Pro&15.54&0.12&4.55&0.19&8.04&0.04&5.51&0.04&1.17&0.19&95.44&0.56&11.50&0.16\\
Owl2.1&23.33&0.25&9.43&0.11&27.3&0.21&18.32&0.33&18.23&0.12&96.76&0.61&22.20&0.21\\
Qwen3-VL&34.44&0.45&29.39&0.23&33.33&0.44&32.22&0.37&29.99&0.40&84.9&0.57&22.20&0.17\\
Gemma-3&50.01&0.61&46.60&0.43&44.44&0.60&42.68&0.49&39.91&0.49&98.0&0.76&41.86&0.47\\
\rowcolor{blue!15}+A-ViT~\cite{avit}
&50.37&0.64&46.60&0.44&45.52&0.63&42.86&0.49&39.91&0.51&98.0&0.77&42.59&0.49\\
\rowcolor{blue!15}+DynamicViT~\cite{dynamicvit}
&50.94&0.66&46.60&0.47&45.83&0.65&43.76&0.49&39.91&0.54&98.04&0.79&43.66&0.51\\
\rowcolor{gray!15}\textbf{+SCEP} (Ours)
&\textbf{51.23}&\textbf{0.67}&\textbf{48.43}&\textbf{0.55}&\textbf{47.32}&\textbf{0.67}&\textbf{46.23}&\textbf{0.43}&\textbf{40.21}&\textbf{0.51}&\textbf{98.06}&\textbf{0.80}&\textbf{54.44}&\textbf{0.53}\\
\bottomrule
\end{tabular}
}
\label{tab::TableI}
\end{table*}

\subsection{Residual Noise Anomaly Score Computation}

Complementary to frequency statistics, deepfake-induced residual-noise artifacts, often concentrated near blended boundaries, are characterized by jointly modeling a local residual mismatch and a global robust deviation. 

For each patch $p_{r,c}$, we apply Gaussian smoothing to obtain a low-pass patch $p^{\mathrm{blur}}_{r,c}$ and compute the residual map $R_{r,c}=|p_{r,c}-p^{\mathrm{blur}}_{r,c}|$, which isolates the attenuated high-frequency components and provides a complementary signal for residual-noise artifact modeling.
The residual energy $E^{\mathrm{res}}_{r,c}$ is then computed as
\begin{equation}
E^{\mathrm{res}}_{r,c}=\frac{1}{|p_{r,c}|}\sum_{x\in p_{r,c}} R(x).
\end{equation}

The noise anomaly score $\mathcal{A}^{\mathrm{noise}}_{r,c}$ is defined as the normalized residual energy $E^{\mathrm{res}}_{r,c}$, 
using the median and median absolute deviation (MAD) as follows:
\begin{equation}
\mathcal{A}^{\mathrm{noise}}_{r,c}=
\frac{E^{\mathrm{res}}_{r,c}-\mathrm{median}(E^{\mathrm{res}})}
{\mathrm{MAD}(E^{\mathrm{res}})+\epsilon},
\end{equation}
where $\mathrm{median}(E^{\mathrm{res}})$ is the median of $E^{\mathrm{res}}_{r,c}$ over all patches, $\mathrm{MAD}(E^{\mathrm{res}})$ is the median absolute deviation of $E^{\mathrm{res}}$, and $\epsilon$ is a small constant for numerical stability.

The suspiciousness score $\mathcal{S}_{r,c}$ for each patch $p_{r,c}$ is defined as 
\begin{equation}
\mathcal{S}_{r,c}
=
\mathcal{A}^{\mathrm{sem}}_{r,c}
+
\alpha\!\left(\mathcal{A}^{\mathrm{freq}}_{r,c}+\mathcal{A}^{\mathrm{noise}}_{r,c}\right),
\label{eq:patch_score_fusion}
\end{equation}
where $\alpha$ balances the contribution of the low-level cues relative to the semantic discrepancy.
Patches are ranked within each semantic cluster based on $\mathcal{S}_{r,c}$ to maintain semantic coherence during evidence mining.
For each cluster ${\mathrm{{C}_k}}$, the top-$k_1$ patches $\mathrm{\hat{S}_{k}}$ are selected by the fused score:
\begin{equation}
\hat{\mathcal{S}}_{k}
=
\operatorname{Top}\text{-}k_1\!\Big(\{\mathcal{S}_{r,c}\mid p_{r,c}\subseteq \mathcal{C}_k\}\Big).
\end{equation}


To reduce redundant selections from spatially adjacent patches, non-maximum suppression $NMS(\cdot)$ is applied to $\mathrm{\hat{S}_{k}}$ using patch boxes on the image grid~\cite{neubeck2006efficient}, yielding the final suspicious token set $\mathcal{S}_k$:
\begin{equation}
\mathcal{S}_k=\mathrm{NMS}\big(\mathrm{\hat{S}_{k}};\tau\big),
\end{equation}
where \(\tau\) is the overlap threshold. 

The resulting $\mathcal{S}_k$ are finally aggregated as a compact, manipulation-revealing evidence pack to condition the frozen LVLM for the IDD decision.

\section{EXPERIMENT}

\begin{figure*}[t!]

\begin{center}
\includegraphics[width=1.0\linewidth]{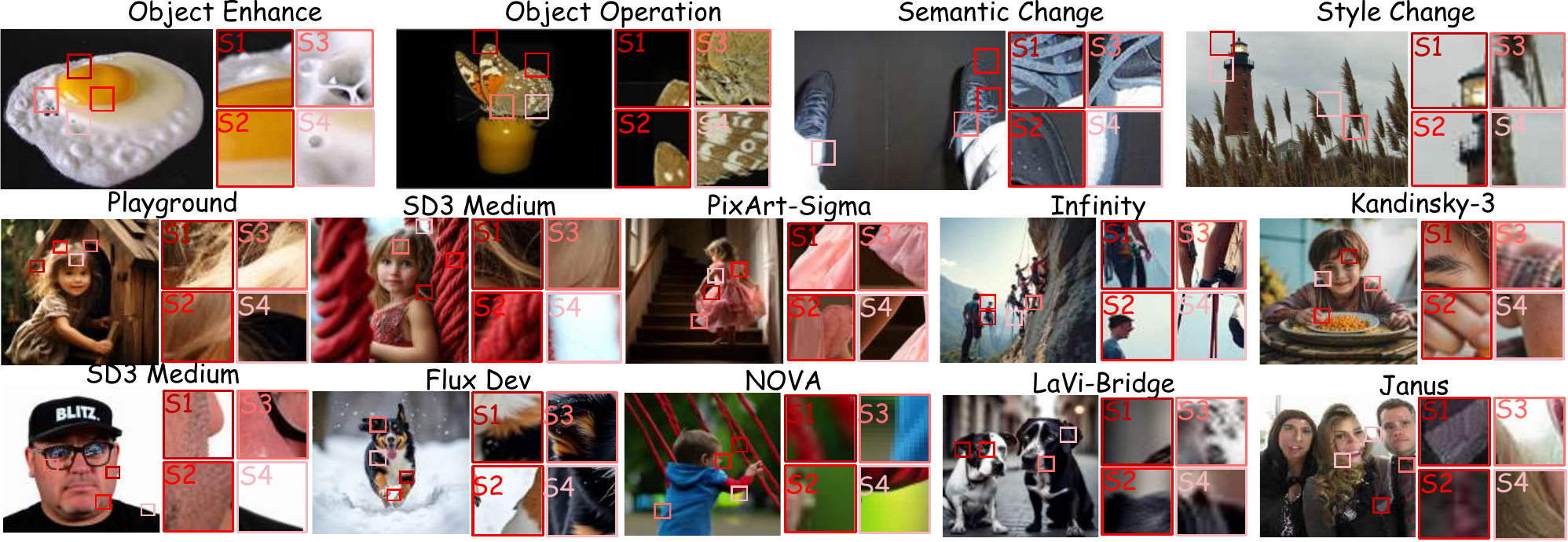}
\vspace{-1.5em}
\caption{
Representative cases from DFBench with the evidence patches selected by our method.
The first row shows AI-edited subsets, while the second and third rows show AI-generated subsets.
S1-S4 denote the top evidence patches sampled from distinct semantic clusters and ranked by the fused anomaly score.
S1-S2 typically localize the dominant manipulated regions, while S3-S4 provide auxiliary evidence capturing subtle or secondary manipulation cues.
}
\label{fig:compareaiedit}
\end{center}
\vspace{-1.5em}
\end{figure*}
\subsection{Experimental Settings}\label{DAT}
\subsubsection{\bfseries Datasets}
We evaluate SCEP on DFBench~\cite{wang2025dfbench}, a large-scale IDD benchmark with three partitions: \textit{Real}, \textit{AI-edited}, and \textit{AI-generated}. 
\textit{AI-generated} images typically contain the richest fine-grained spatial details, whereas \textit{AI-edited} samples exhibit intermediate characteristics, preserving authentic regions while introducing localized manipulations.

\subsubsection{\bfseries Implementaion Details}
We evaluate SCEP in a training-free IDD setting, where each image is classified as Real or Fake. Results are reported in Tables~\ref{tab::TableI}, \ref{tab::TableII}, and \ref{tab::TableIII} using Accuracy (Acc\%) and F1-score (F1) over the AI-generated and Real-image partitions and their corresponding datasets, as well as the AI-edited partition spanning diverse manipulation types, together with Overall performance.
Accordingly, we instantiate our framework on multiple frozen LVLMs, including Kimi-VL-16B~\cite{kimivl}, Gemma-3-12B~\cite{team2025gemma}, MiniCPM-V-8B~\cite{yu2025minicpm}, Janus-Pro-7B~\cite{Janus-pro}, Owl2.1-7B~\cite{mplug-owl2}, and  Qwen3-VL-8B~\cite{yang2025qwen3}, with no task-specific fine-tuning.
We further compare our approach with A-ViT~\cite{avit} and DynamicViT~\cite{dynamicvit} by evaluating them using their best-performing backbone model.
In the Real-image partitions, following DFBench, we report only Accuracy (Acc\%).
Specifically, we vary $\alpha \in \{0,0.3,0.7,0.9\}$ to examine how the frequency and noise anomaly terms in Eq.~\ref{eq:patch_score_fusion} affect detection performance across different partitions and manipulation types.

We additionally report the inference latency of our framework, with all experiments conducted on an NVIDIA A6000 server equipped with an Intel(R) Xeon(R) Gold 6330 CPU.

\subsection{\bfseries Performance Analysis}

From Table~\ref{tab:merged}, on the AI-generated subsets, SCEP achieves its strongest gains on Gemma-3, getting the overall accuracy of 54.44\% and the F1 score of 0.53.
On the AI-edited subsets, the largest improvement is observed with Qwen3-VL, where SCEP gets the overall accuracy of 61.91\% and the F1 score of 0.58.
The larger gains on AI-generated and AI-edited images suggest that global semantic prompting alone misses fine-grained synthesis artifacts and localized edits, while SCEP’s fused semantic, frequency, and noise cues provide more discriminative evidence for IDD without task-specific fine-tuning.
On the Real-image subsets, SCEP preserves strong overall performance without noticeable degradation, suggesting that the frequency and noise cues are non-intrusive on authentic images.

\begin{table}[t!]
\centering
\setlength{\tabcolsep}{6pt}          
\renewcommand{\arraystretch}{1.12}   
\small
\caption{SCEP on DFBench real-image Datasets.}
\vspace{-0.6em}

\resizebox{\linewidth}{!}{
\begin{tabular}{@{} C{2.2cm} *{6}{c} @{}}
\toprule
\multicolumn{1}{c}{\textbf{Methods}}
& \textbf{LIVE~\cite{LIVE}}
& \textbf{CSIQ~\cite{CSIQ}}
& \textbf{TID2013~\cite{TID2013}}
& \textbf{KADID~]\cite{kADID-10k}}
& \textbf{KonIQ-10k~\cite{KonIQ-10k}}
& \textbf{Overall} \\
\midrule
Kimi-VL     & 66.32 & 53.12 & 75.21 & 60.03 & 85.32 & 69.39 \\
Gemma-3     & 71.21 & 75.32 & 67.21 & 57.11 & 69.15 & 68.33 \\
MiniCPM-V   & 70.00 & 79.12 & 74.21 & 66.60 & 71.21 & 72.56 \\
Janus-Pro   & 61.21 & 54.21 & 55.31 & 58.03 & 66.13 & 59.68 \\
Owl2.1      & 77.12 & 68.99 & 70.02 & 77.43 & 71.71 & 72.44 \\
Qwen3-VL     
& 87.81 & 87.76 & 84.31 & 88.21 & 79.38 & 84.87 \\
\rowcolor{blue!15}+A-ViT
& 89.65 & 87.34 & 87.45 & 88.01 & 84.11 & 86.71 \\
\rowcolor{blue!15}+DynamicViT
& 90.65 & 88.54 & 85.56 & 88.37 & 83.67 & 86.75 \\
\rowcolor{gray!15}
\textbf{+SECP} (Ours)
& \textbf{91.74} & \textbf{90.03} &\textbf{ 88.32} & \textbf{89.92 }& \textbf{85.32} & \textbf{88.09} \\
\bottomrule
\end{tabular}
}
\label{tab::TableII}
\end{table}

\begin{table}[t]
\label{tab::TableIV}
\centering
\caption{Performance with Varying $\alpha$ on AI-Edited.}
\label{tab:GSTAS-alpha}
\vspace{-0.45em}

\setlength{\tabcolsep}{3.5pt}        
\renewcommand{\arraystretch}{1.04}   
\small

\resizebox{\linewidth}{!}{
\begin{tabular}{@{} C{1.35cm} *{12}{c} @{}}
\toprule
\multirow{2}{*}{\centering\textbf{Setting}} &
\multicolumn{2}{c}{\textbf{Object Enhance}} &
\multicolumn{2}{c}{\textbf{Object Operation}} &
\multicolumn{2}{c}{\textbf{Semantic Change}} &
\multicolumn{2}{c}{\textbf{Style Change}} &
\multicolumn{2}{c}{\textbf{Real Source}} &
\multicolumn{2}{c}{\textbf{Overall}} \\
\cmidrule(lr){2-3}\cmidrule(lr){4-5}\cmidrule(lr){6-7}\cmidrule(lr){8-9}\cmidrule(lr){10-11}\cmidrule(lr){12-13}
& \textbf{Acc\%} & \textbf{F1}
& \textbf{Acc\%} & \textbf{F1}
& \textbf{Acc\%} & \textbf{F1}
& \textbf{Acc\%} & \textbf{F1}
& \textbf{Acc\%} & \textbf{F1}
& \textbf{Acc\%} & \textbf{F1} \\
\midrule
$\alpha=0$   & 48.93 & 0.38 & 41.28 & 0.34 & 48.34 & 0.40 & 49.87 & 0.48 & 83.74 & 0.63 & 68.03 & 0.45 \\
$\alpha=0.3$ & 50.67 & 0.45 & 50.43 & 0.52 & 49.93 & 0.43 & 50.13 & 0.53 & 88.34 & 0.68 & 57.90 & 0.52 \\
\rowcolor{gray!15} {$\alpha=0.7$}
&\textbf{ 57.32} & \textbf{0.56} & \textbf{50.01 }& \textbf{0.59} & \textbf{52.21 }& \textbf{0.47 }& \textbf{60.00} &\textbf{ 0.61} & \textbf{90.01} & \textbf{0.70} & \textbf{61.91 }&\textbf{ 0.59} \\
$\alpha=0.9$ & 54.07 & 0.53 & 52.94 & 0.50 & 51.74 & 0.45 & 51.60 & 0.49 & 79.89 & 0.58 & 58.05 & 0.51 \\
\bottomrule
\end{tabular}
}
\end{table}
\vspace{-0.3em}
\subsection{Case Study on Evidence Patch Mining}
Fig.~\ref{fig:compareaiedit} provides qualitative evidence that complements the quantitative results in Tables~\ref{tab::TableI}~\ref{tab::TableII}~\ref{tab::TableIII}.
The AI-edited cases shows that the evidence patches for Object Enhance and Object Operation predominantly localize to manipulated regions, capturing abrupt texture discontinuities and localized high-frequency inconsistencies, rather than semantically correct yet uninformative backgrounds.
For Style Change, the patches consistently attend to regions with style-induced spectral shifts, indicating that low-level anomalies remain reliable complementary cues when semantic consistency is largely preserved.
In AI-generated subjects , the evidence patches tend to concentrate on generation-specific artifacts, such as globally over-smoothed textures, unnatural high-frequency statistics, and inconsistent fine-grained details.
These observations verify that SCEP consistently retrieves semantically grounded yet anomaly-sensitive evidence, thereby improving IDD detection accuracy across both localized edits and  synthesized images.

\begin{table}[t]

\centering
\footnotesize
\setlength{\tabcolsep}{2.2pt}
\renewcommand{\arraystretch}{1.06}

\caption{Performance on DFBench On AI-Edited Subsets.}
\label{tab:merged}
\vspace{-0.6em}

\resizebox{\linewidth}{!}{
\begin{tabular}{@{} C{2.05cm} *{12}{c} @{}}
\toprule
\multirow{2}{*}{\textbf{Methods}}
& \multicolumn{2}{c}{\textbf{Object Enhance}}
& \multicolumn{2}{c}{\textbf{Object Operation}}
& \multicolumn{2}{c}{\textbf{Semantic Change}}
& \multicolumn{2}{c}{\textbf{Style Change}}
& \multicolumn{2}{c}{\textbf{Real Source}}
& \multicolumn{2}{c}{\textbf{Overall}} \\
\cmidrule(lr){2-3}\cmidrule(lr){4-5}\cmidrule(lr){6-7}\cmidrule(lr){8-9}\cmidrule(lr){10-11}\cmidrule(lr){12-13}
 & \textbf{Acc}\% &\textbf{ F1}
& \textbf{Acc}\% &\textbf{ F1}
& \textbf{Acc\% }& \textbf{F1}
& \textbf{Acc\% }& \textbf{F1}
& \textbf{Acc\%} & \textbf{F1}
& \textbf{Acc\% }& \textbf{F1 }\\
\midrule

Kimi-VL     &11.21&0.19&13.21&0.23&21.21&0.34&19.32&0.16&83.32&0.64&29.65&0.31\\
Gemma-3     &21.34&0.21&23.32&0.21&15.65&0.12&19.92&0.19&85.54&0.61&33.15&0.27\\
MiniCPM-V   &13.32&0.23&19.03&0.29&28.32&0.34&33.32&0.28&81.21&0.52&35.04&0.33\\
Janus-Pro   &16.21&0.17&23.12&0.21&23.32&0.32&28.12&0.21&87.77&0.58&35.71&0.30\\
Owl2.1      &36.32&0.32&23.45&0.21&27.32&0.31&30.02&0.25&82.32&0.53&39.89&0.32\\
Qwen3-VL    
&53.43&0.49&45.54&0.54&49.09&0.40&57.54&0.51&87.66&0.56&58.65&0.50\\

\rowcolor{blue!15} +A-ViT
&55.77&0.50&46.32&0.55&50.14&0.41&57.98&0.53&87.78&0.61&59.59&0.52\\

\rowcolor{blue!15} +DynamicViT
&55.97&0.51&48.82&0.57&51.89&0.47&58.01&0.57&88.81&0.62&60.69&0.54\\

\rowcolor{gray!15} \textbf{+SECP} (Ours)
&\textbf{57.32}&\textbf{0.56}&\textbf{50.01}&\textbf{0.59}&\textbf{52.21}&\textbf{0.47}&\textbf{60.00}&\textbf{0.61}&\textbf{90.01}&\textbf{0.70}&\textbf{61.91}&\textbf{0.58}\\
\addlinespace[0.10em]
\bottomrule
\end{tabular}
}
\label{tab::TableIII}
\end{table}
\subsection{Inference Latency}
As shown in Table~\ref{tab::infertime}, SCEP consistently reduces inference latency across all frozen LVLM, delivering speedups ranging about 1.58 times.
\begin{table}[t]
\centering
\caption{Per-image inference time (s) of frozen LVLMs with and without SCEP.}
\label{tab:inference_time}
\vspace{-0.45em}

\setlength{\tabcolsep}{4.2pt}        
\renewcommand{\arraystretch}{1.04}   
\small

\resizebox{\linewidth}{!}{
\begin{tabular}{@{} C{2.4cm} *{5}{c} @{}}
\toprule
\multirow{2}{*}[-0.3em]{\centering\textbf{Setting}} &
\multicolumn{5}{c}{\textbf{LVLM inference time (s)}} \\
\cmidrule(lr){2-6}
& \textbf{Kimi-VL-16B} & \textbf{MiniCPM-V-8B} & \textbf{Janus-Pro-7B} & \textbf{Owl2.1-7B} & \textbf{Qwen3-VL-8B} \\
\midrule
Vanilla              & 33.17 & 23.76 &  26.39  & 24.15  & 21.44\\
\rowcolor{gray!15}
\textbf{+SCEP }(Ours) & 28.36 & 16.72 & 19.34  & 20.89  & 13.56\\

\bottomrule
\end{tabular}
}
\label{tab::infertime}
\end{table}
\section{CONCLUSIONS}
We propose SCEP, a training-free LVLM-based framework for IDD that moves beyond costly fine-tuning and brittle whole-image inference.
SCEP mines a compact evidence set of manipulation-revealing visual tokens via patch clustering and fused anomaly scoring.
In particular, frequency cues capture spectral irregularities such as abnormal high-frequency energy, while residual-noise cues expose boundary-localized inconsistencies induced by blending or synthesis.
Experiments on DFBench demonstrate that SCEP consistently improves IDD accuracy and robustness across diverse manipulation types while remaining training-free and efficient.
\bibliographystyle{IEEEbib}
\bibliography{icme2026references}

\end{document}